\documentclass[10pt,twocolumn,letterpaper]{article}

\usepackage[pagenumbers]{cvpr} 

\usepackage{graphicx}
\usepackage{amsmath}
\usepackage{amssymb}
\usepackage{booktabs}

\usepackage[pagebackref,breaklinks,colorlinks]{hyperref}

\usepackage[capitalize]{cleveref}
\crefname{section}{Sec.}{Secs.}
\Crefname{section}{Section}{Sections}
\Crefname{table}{Table}{Tables}
\crefname{table}{Tab.}{Tabs.}

\def\cvprPaperID{*****} 
\def\confName{CVPR}
\def\confYear{2022}

\begin{document}

\title{Cutting-Edge Techniques for Depth Map Super-Resolution}

\author{Ryan Peterson\\
University of Texas at Dallas\\
{\tt\small ryan.peterson@utdallas.edu}
\and
Josiah Smith\thanks{Both authors have equal contribution for this work.}\\
University of Texas at Dallas\\
{\tt\small josiah.smith@utdallas.edu}
}

\maketitle






\begin{abstract}
To overcome hardware limitations in commercially available depth sensors which result in low-resolution depth maps, depth map super-resolution (DMSR) is a practical and valuable computer vision task. 
DMSR requires upscaling a low-resolution (LR) depth map into a high-resolution (HR) space. 
Joint image filtering for DMSR has been applied using spatially-invariant and spatially-variant convolutional neural network (CNN) approaches. 
In this project, we propose a novel joint image filtering DMSR algorithm using a Swin transformer architecture \cite{liu2021swin,liu2021swinv2}.
Furthermore, we introduce a Nonlinear Activation Free (NAF) network based on a conventional CNN model used in cutting-edge image restoration applications and compare the performance of the techniques.
The proposed algorithms are validated through numerical studies and visual examples demonstrating improvements to state-of-the-art performance while maintaining competitive computation time for noisy depth map super-resolution. 
\end{abstract}

\section{Introduction}
\label{sec:intro}
Depth estimation is a key problem in computer vision to enable scene understanding, navigation, segmentation, etc. 
However, depth sensors are typically low-resolution compared to their optical counterparts. 
Hence, depth map super-resolution (DMSR) is of interest for practical applications of depth sensors. 
We examined prior work \cite{he2021towards} on DMSR to develop a thorough intuition of the problem-space and built upon state-of-the-art computer vision and deep learning techniques \cite{liu2021swin,liu2021swinv2,dosovitskiy2020image}. 
In current state-of-the-art techniques, researchers primarily rely on convolutional neural networks (CNNs) as the mechanism for detecting and transferring features from a guidance image to a low resolution depth image \cite{kim2021deformable} to perform super-resolution. 
\begin{figure}[t]
    \centering
    \includegraphics[width=0.3\textwidth]{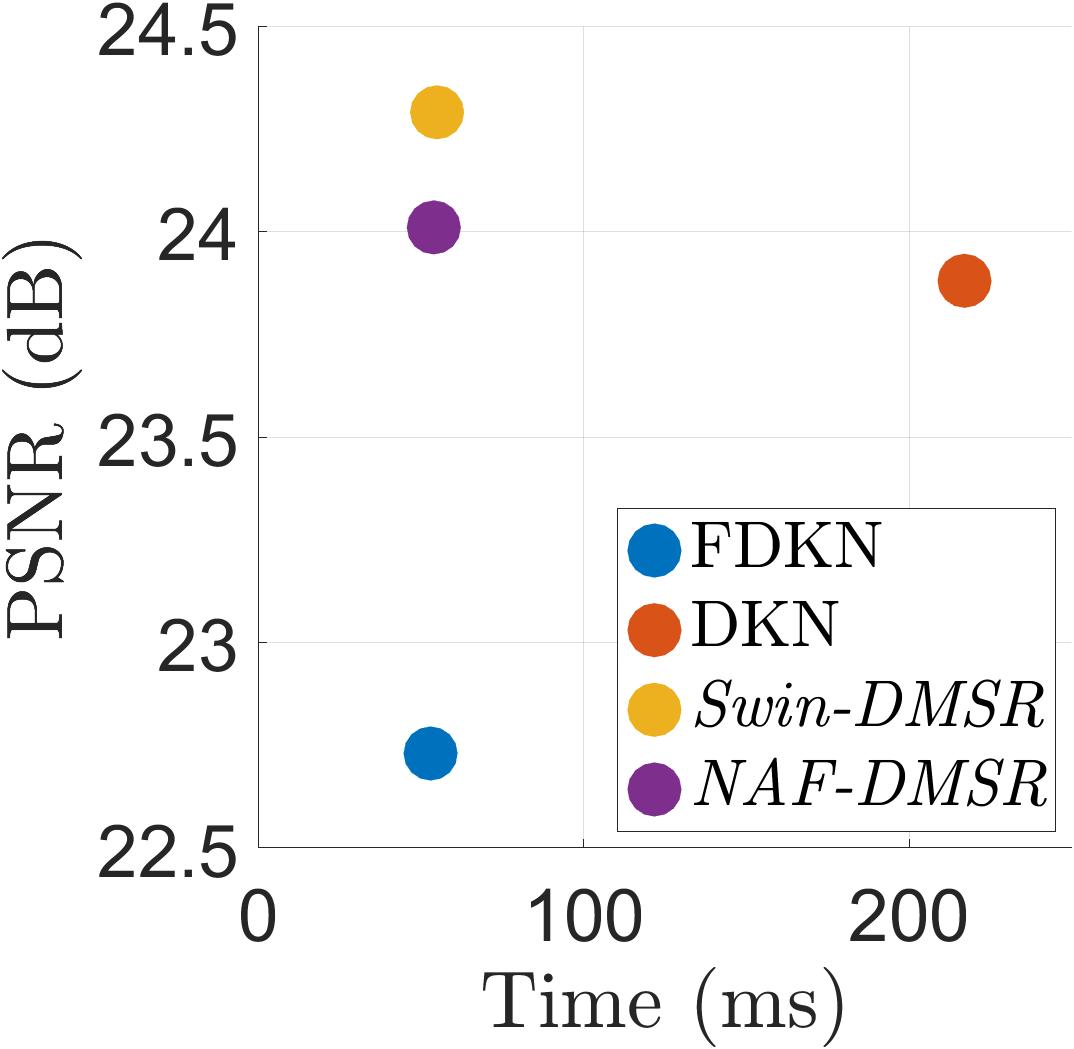}
    \caption{NYU V2 Depth Map Super-Resolution x8 results for current state-of-the-art algorithms (FDKN and DKN \cite{kim2021deformable}) compared to the proposed algorithms (\textit{Swin-DMSR} and \textit{NAF-DMSR}) in terms of numerical performance (PSNR) and computation time. We demonstrate that a swin- or NAF-based architecture can achieve superior super-resolution performance to DKN with comparable computational speed to the FDKN.}
    \label{fig:sota}
\end{figure}
However, as the attention mechanism, an extension of recurrent neural networks (RNNs), has recently transformed the arenas of natural language processing (NLP) \cite{vaswani2017attention} and computer vision \cite{dosovitskiy2020image}, we investigate an attention-based solution to DMSR.
Recently, a shifted window (Swin) Transformer architecture \cite{liu2021swin,liu2021swinv2} has emerged as a competitive alternative to the vision transformer \cite{dosovitskiy2020image,smith2022ffh_vit} achieving lower computational complexity and improved scalability.
The Swin transformer architecture has been applied to an array of applications \cite{liang2021swinir,xie2021self,cao2021swin} demonstrating state-of-the-art performance with minimal computational expense. 
We propose a novel DMSR technique using the Swin transformer to achieve higher performance at a lower computational cost.

Additionally, recent techniques have shown significant improvements at reduced computational costs in networks that have been simplified to exclude nonlinear activation functions such as ReLU \cite{chen2022nafnet}. 
We also propose and evaluate networks using this technique, comparing the results to both traditional CNN approaches and our Swin-based networks.

Building off the work in \cite{liu2021swin,chen2022nafnet}, we detail two networks using a joint image filtering technique for efficient, high-fidelity DMSR.
The algorithms developed in this project achieve state-of-the-art DMSR performance with competitive computational efficiency, as shown in Fig. \ref{fig:sota}. 

The remainder of the report is formatted as follows. 
In Section \ref{sec:background}, we detail the joint image filtering technique and relevant advances in DMSR from the deep learning community. 
Section \ref{sec:methods} overviews the two novel learning-based architectures proposed in this project using the swin transformer and nonlinear activation free (NAF) approach as the backbone for DMSR. 
The proposed algorithms are validated by numerical studies and visual examples in Section \ref{sec:experiments} demonstrating state-of-the-art DMSR performance with high computational efficiency, followed finally by conclusions in Section \ref{sec:conclusion}. 

\section{Related Work}
\label{sec:background}
In this section, we examine the context of our approach in the existing literature for learning-based DMSR and image restoration, starting with preliminaries on joint image filtering. 

\begin{figure*}[t]
    \centering
    \includegraphics[width=\textwidth]{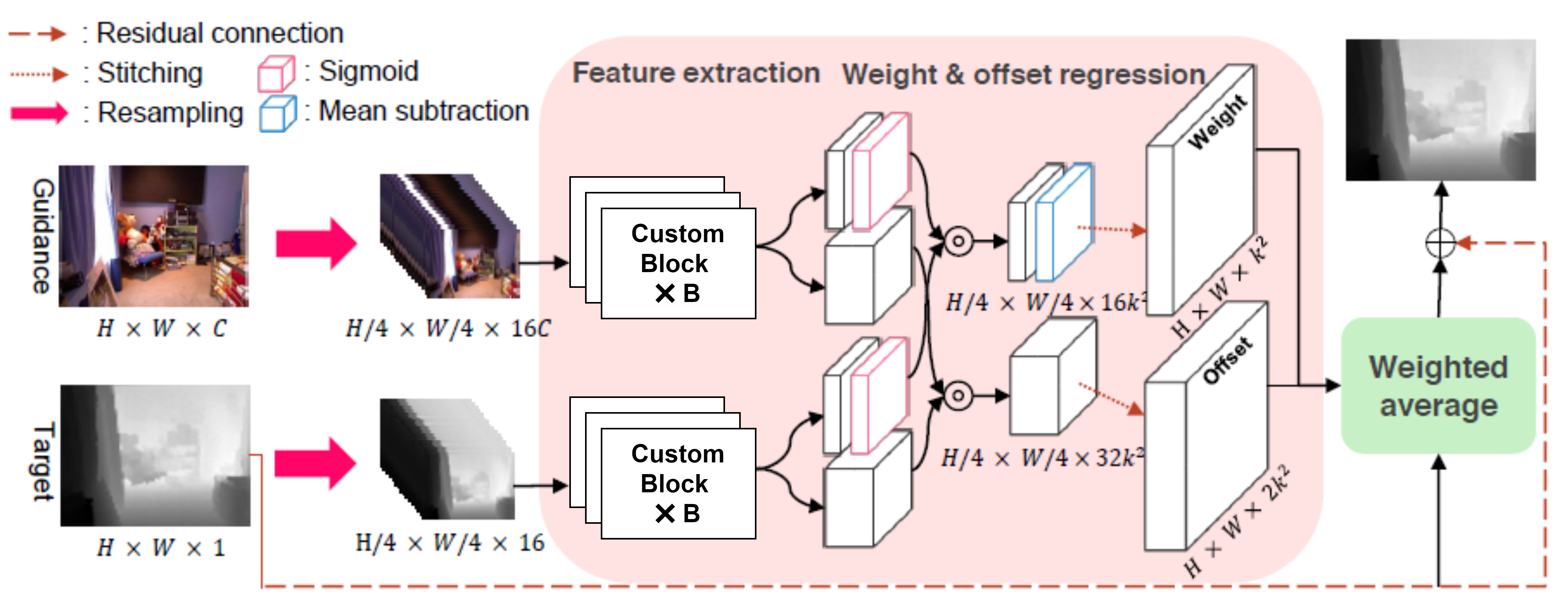}
    \caption{Proposed architecture of the algorithms developed in this project. Guidance and target images are first resampled. The resampled images are then fed into $B$ blocks, which vary by the specific approach (Swin or NAF), before two convolution layers are applied to yield weight and bias tensors. The tensors from the guidance and target images are combined and stitched into a tensor with the same dimensions as the original and applied to the target image after a weighted average \cite{he2021towards}.}
    \label{fig:architecture}
\end{figure*}

\subsection{Joint Image Filtering} 
DMSR is often approached using a joint image filtering technique. 
Joint image filtering uses two different components: (1) a guidance image and (2) a target image. The guidance image has structural components such as texture or a higher fidelity that is transferred to the target image. 
Joint image filtering has been applied to a vast array of computer vision and graphics processing tasks, such as cross-modality image restoration \cite{shen2015mutual}, texture removal \cite{zhang2014rolling}, scale-space filtering \cite{ham2017robust}, dense correspondence \cite{hosni2012fast}, semantic segmentation \cite{barron2016fast}, and depth map up-sampling \cite{kim2021deformable,ham2017robust}. 
In this project, we focus on joint image filtering for DMSR.

The joint image filtering operation can be formulated as 
\begin{equation}
    \label{eq:jif_1}
    \hat{f}_\mathbf{p} = \sum_{\mathbf{q} \in \mathcal{N}(\mathbf{p})} \mathbf{W}_{\mathbf{pq}}(f,g) f_\mathbf{q},
\end{equation}
where $\mathbf{p} = (x,y)$ is a position on the output image $\hat{f}$, which is the weighted average of the target image $f$ and guidance image $g$ expressing the weighting kernel $\mathbf{W}$. 
$\mathcal{N}(\mathbf{p})$ is a set of neighbors near the position $\mathbf{p}$ on a discrete Cartesian grid. 
Furthermore, the kernel $\mathbf{W}$ is normalized such that

\begin{equation}
    \label{eq:jif_W_norm}
    \sum_{\mathbf{q} \in \mathcal{N}(\mathbf{p})} \mathbf{W}_{\mathbf{pq}}(f,g) = 1.
\end{equation}

Whereas classical techniques for joint image filtering handcraft kernels $\mathbf{W}$ and neighbors $\mathcal{N}$, current state-of-the-art approaches to joint image filtering use CNNs as the primary method for detecting and transferring desired features \cite{kim2021deformable}.
In this project, we employ the joint image filtering approach in (\ref{eq:jif_1}) for DMSR using recent developments in image restoration. 

\subsection{Image Restoration}
Image restoration is a task closely related to joint image filtering. 
The primary distinguishing factor of image restoration is that it seeks to improve the quality of images without any reference image. 
The exact task can be very broad, ranging from deblurring to noise removal to super-resolution \cite{chen2022nafnet}. 

The range of applications for image restoration yields itself to rapid development and, as a result, many adjacent tasks such as joint image filtering are not utilizing the most recent developments.
State-of-the-art improvements to image restoration have ranged from transformer-based architectures \cite{liang2021swinir} to Nonlinear Activation Free (NAF) networks \cite{chen2022nafnet} in the past two years. 
In this project, we leverage these advancements in combination with architectures developed for the joint image filtering task to improve the accuracy and speed of DMSR networks.

\subsection{Depth Map Super-Resolution} 
The improvements proposed in this paper are twofold in the area of DMSR. 
Firstly, the accuracy of super-resolution is the goal for any super-resolution task and state-of-the-art approaches are primarily based on this metric \cite{kim2021deformable}. 
However, the second area of focus is model throughput. 
The most recent advancements in DMSR not only boast improved accuracy of the model output but also decreased computation time \cite{zhong2021high}. 
Although Vision Transformer (ViT) frameworks have yielded improved numerical and computational performance compared to convolution-based approaches \cite{cao2021swin,liu2021swin,liu2021swinv2,liang2021swinir,smith2022ffh_vit}, transformer-based architectures for DMSR have not been explored in the existing literature.
By introducing the swin transformer in place of CNNs, we hope to also improve upon the accuracy and run time due to its inherent reduction in computational complexity paired with its improvements in inference capabilities \cite{liu2021swin,liang2021swinir}. 

Furthermore, we employ recent developments for CNN-based image restoration using Nonlinear Activation Free (NAF) networks to develop novel DMSR algorithms that replace nonlinear activations with multiplication and see improved performance \cite{chen2022nafnet}. 

\begin{figure}[hb]
     \centering
     \begin{subfigure}[b]{0.45\textwidth}
         \centering
         \includegraphics[width=\textwidth]{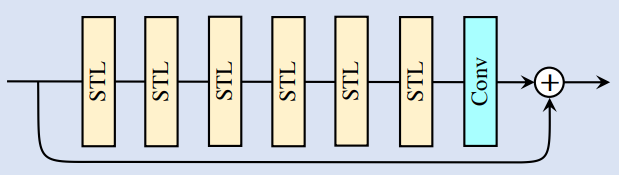}
         \caption{A residual swin transformer block (RSTB) containing several swin transformer layers.}
         \label{fig:rstb}
     \end{subfigure}
     \hfill
     \begin{subfigure}[b]{0.45\textwidth}
         \centering
         \includegraphics[width=\textwidth]{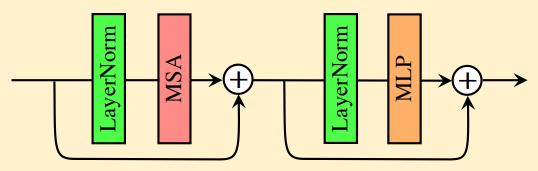}
         \caption{A swin transformer layer (STL) with multi-head self attention (MSA) and multi-layer perceptron (MLP) modules.}
         \label{fig:stl}
     \end{subfigure}
        \caption{Residual swin transformer block (RSTB) and swin transformer layer (STL) used in the proposed \textit{Swin-DMSR} algorithm \cite{liang2021swinir}.}
        \label{fig:swin_blocks}
\end{figure}

\section{Method}
\label{sec:methods}
This section details the architectures and methods of our proposed implementations. 

\subsection{Overview} 
\begin{figure}[h]
  \centering
  \begin{tabular}[c]{cc}
    \subfloat[(top) SimpleGate (SG) and (bottom) simplified channel attention (SCA) architectures.]{\label{fig:sg_sca}%
      \begin{tabular}[b]{c}
        \includegraphics[width=0.3\textwidth]{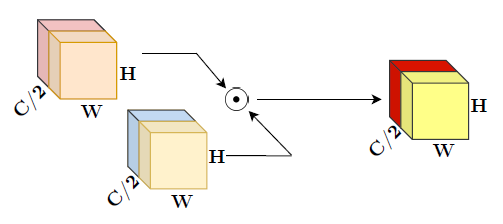}\\[2mm]
        \includegraphics[width=0.3\textwidth]{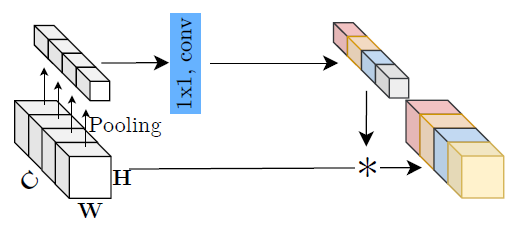}
      \end{tabular}}
    &
    \begin{tabular}[b]{c}
      \subfloat[Nonlinear Activation Free (NAF) block.]{\label{fig:nafblock}%
        \includegraphics[width=0.08\textwidth]{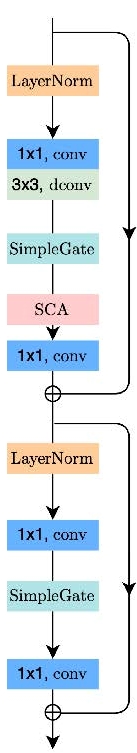}}
    \end{tabular}
  \end{tabular}
  \caption{SimpleGate (SG), simplified self attention (SCA) module, and nonlinear activation free (NAF) block used in the proposed \textit{NAF-DMSR} super-resolution architecture \cite{chen2022nafnet}.}
  \label{fig:naf}
\end{figure}

Fig. \ref{fig:architecture} details the architecture of the  network. 
Our approach takes a guidance and target image as input where the guidance is a high-resolution (HR) RGB image and the target is an up-sampled low resolution (LR) depth image using bicubic interpolation. 
The model then outputs the super-resolution (SR) depth map of the scene. 
The architecture follows a similar structure to the deformable kernel network (DKN) proposed in \cite{kim2021deformable}. 
However, instead of relying on sequential CNNs to perform feature extraction, our novel architecture applies techniques from state-of-the-art advancements in the image restoration field. 
We experiment with both a swin transformer-based model and a NAF network. 
Our novel architectures apply a two-path network and combine those results using weighted averaging to incorporate the features into the original depth map. 
Our swin network employs a similar approach to the swin image restoration (SwinIR) architecture detailed in \cite{liang2021swinir} by using residual swin transformer blocks (RSTBs) which do not downsample the input image as shown in Fig. \ref{fig:rstb}. 
Our NAF-based algorithm utilizes network blocks introduced in \cite{chen2022nafnet}, as show in Fig. \ref{fig:naf}.

\subsection{Design of Proposed Network Architectures} 
This section details the workflow of the networks proposed in this project, which is shown in general form Fig. \ref{fig:architecture}.
Guidance and target images are first resampled into 16 images of lower resolution.
The resampled images are then used as input to parallel networks for feature extraction.
Each of the parallel networks consists of $B$ custom blocks using either the RSTB block, as shown in Fig. \ref{fig:rstb}, or the NAF block, as shown in Fig. \ref{fig:nafblock}. 

\subsubsection{Swin-DMSR Network}
First, we design a network called \textit{Swin-DMSR} using the swin transformer as the backbone for DMSR replacing convolution \cite{liu2021swin,liang2021swinir,liu2021swinv2}. 
The RSTB layer consists of $B$ sequential swin transformer layers (STLs), each of which consists of a residual structure using layer norms (LNs) in conjunction with the multi-head self-attention (MSA) mechanism \cite{dosovitskiy2020image} or a simple multi-layer perceptron (MLP) network, as shown in Fig. \ref{fig:stl}. 
After an empirical study, our implementation uses $B=4$ RSTB layers. 
The output of the \textit{Swin-DMSR} network is used in a final convolutional layer to produce weight and offset tensors for computation of (\ref{eq:jif_1}). 
The weight and offset tensors are used later for joint image filtering to produce the super-resolution image. 

\subsubsection{NAF-DMSR Network}
The second proposed network uses a recent technique boasting state-of-the-art performance for image restoration known as nonlinear activation free network (NAFNet) \cite{chen2022nafnet}. 
Starting with the Gaussian Error Linear Unit (GELU), the general form of many non-linear functions can be expressed as a Gated Linear Unit. 
The GELU is given by \cite{hendrycks2016gelu}
\begin{equation}
    \label{eq:gelu}
    \text{GELU}(\mathbf{x}) = \mathbf{x} \odot \phi(\mathbf{x}),
\end{equation}
where $\phi(\cdot)$ is the cumulative distribution function of the standard normal distribution and $\odot$ denotes the Hadamard or element-wise product.
The GELU can be understood as a smooth adaptation of the Rectified Linear Unit (ReLU) \cite{glorot2011ReLU} with more optimal statistical characteristics at scale, and can be approximated by 
\begin{equation}
    \label{eq:gaussian_cdf_approximation}
    \text{GELU}(\mathbf{x}) \approx 0.5 \mathbf{x} \odot \left( 1 + \tanh \left[ \sqrt{2/\pi} \left( \mathbf{x} + 0.044715 \mathbf{x} \right) \right] \right). 
\end{equation}

From (\ref{eq:gelu}), the GELU is a special case of the GLU, whose general form is given as 
\begin{equation}
    \label{eq:glu}
    \text{GLU}(\mathbf{x},f,g,\sigma) = f(\mathbf{x}) \odot \sigma \left( g(\mathbf{x}) \right), 
\end{equation}
where $f(\cdot)$ and $g(\cdot)$ are linear transformers and $\sigma(\cdot)$ is a nonlinear activation function, e.g. sigmoid. 

In \cite{chen2022nafnet}, the SimpleGate (SG) is proposed as a simple GLU removing the nonlinear activation and using $f$ and $g$ as identity functions yielding
\begin{equation}
    \label{eq:simplegate}
    \text{SimpleGate}(\mathbf{y}, \mathbf{z}) = \mathbf{y} \odot \mathbf{z},
\end{equation}
where $\mathbf{y}$ and $\mathbf{z}$ consist of half of the activation map split along the channel dimension or $\mathbf{x} = [\mathbf{y} \ \mathbf{z}]$, as shown in Fig. \ref{fig:sg_sca}. 
Hence, the input to the SimpleGate, an activation map $\mathbf{x}$ with dimension $H \times W \times C$, is split into two tensors $\mathbf{y}$ and $\mathbf{z}$ of size $H \times W \times C/2$. 
The Hadamard product of those two tensors is the output of the SimpleGate with half the channels as the input. 

As a result, the NAF block, shown in Fig. \ref{fig:nafblock}, uses $1 \times 1$ convolution layers to increase the number of channels by a factor of 2 to accommodate the channel reduction. 
The NAF block also uses a Simplified Channel Attention (SCA) module, as shown in Fig. \ref{fig:sg_sca}, building off of the channel attention module in \cite{hu2018squeeze} with a global receptive field while maintaining computational efficiency. 
More details on the SG and SCA mechanisms are available in \cite{chen2022nafnet}. 

The proposed \textit{NAF-DMSR} network employs $B = 6$ NAF blocks connected in series to produce features used in constructing the weight and offset tensors, similarly to the \textit{Swin-DMSR}. 
After the weight and offset are generated, they are recombined to recover the super-resolution image as follows. 

\subsubsection{Recombination of Weight and Offset to Recover the Super-Resolution Image}
A sigmoid function is applied to the weight tensor and the result of this operation from the target and guidance images are combined by applying element-wise multiplication. 
Mean subtraction is then performed on the combined weight tensors and then finally stitched to produce a tensor with a $H \times W \times k^2$ dimensionality where $H$ and $W$ are the height and width of the original image, respectively, and $k$ is the super-resolution factor.

The offset tensors are combined through a similar process, but there are two key differences. 
(1) The CNN applied to create the offset applies more kernels and results in $2k^2$ channels in the final result. 
(2) No sigmoid or mean subtraction layers are applied to the resulting vector. 
The output of the guidance and target is immediately combined using element-wise multiplication and then stitched into a $H \times W \times 2k^2$ tensor.

\begin{table*}[h]
    \centering
    \begin{tabular}{c || c | c | c | c}
         \hline \hline
              & FDKN \cite{kim2021deformable} & DKN \cite{kim2021deformable} & \textit{Swin-DMSR}  & \textit{NAF-DMSR}  \\
         \hline \hline
         PSNR (dB) & 22.73 & 23.88 & \textbf{24.29} & 24.01 \\
         \hline
         Time (ms) & \textbf{53} & 217 & 55 & 54 \\
         \hline \hline
    \end{tabular}
    \caption{Quantitative comparison of state of the art on noisy depth map super-resolution performance in terms of PSNR on the NYU v2 dataset. The networks in italics are proposed in this project and the best result is given in boldface.}
    \label{tab:rmse_results}
\end{table*}

The authors would like to point out that the DKN \cite{kim2021deformable} employs a different architecture than the FDKN, which is used as a baseline for our novel models. 
We have investigated using the swin and NAF mechanism as the backbone for models using a similar architecture to the DKN, which are included in the provided source code. 
However, numerical performance and efficiency both demonstrated empirical superiority using the simplified FDKN baseline. 
Additionally, the proposed networks achieve state-of-the-art numerical performance while demonstrating comparable efficiency to the FDKN. 

\section{Experiments}
\label{sec:experiments}
In this section, we detail our initial findings applying the proposed network for DMSR. 

\begin{figure*}[h]
    \centering
    \includegraphics[width=\textwidth]{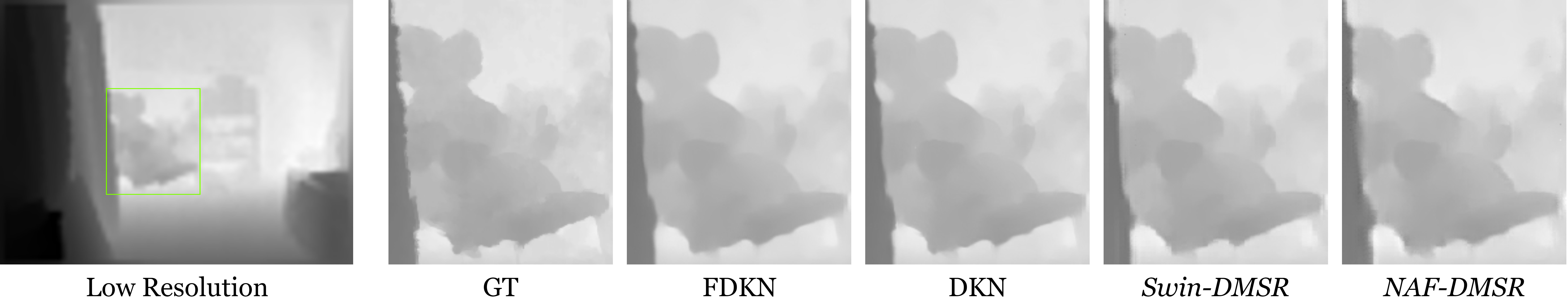}
    \caption{Visual comparison of the super-resolution techniques (FDKN, DKN \cite{kim2021deformable}, \textit{Swin-DMSR}, and \textit{NAF-DMSR}) on a low-resolution image with the corresponding ground-truth (GT) high-resolution image. A portion of the entire image is selected to evaluate the performance of each algorithm on this sample.}
    \label{fig:comparison}
\end{figure*}

\subsection{Datasets and Evaluation Metrics}
To train our networks, we follow the protocol of \cite{li2016nyu,kim2021deformable} using a large set of RGB/D image pairs for DMSR using the structural details from the guidance images for joint image filtering. 
We employ RGB/D image pairs of size $640 \times 480$ from the NYU v2 dataset \cite{silberman2012nyuv2} to train the networks. 
The same training and testing split as used in \cite{li2016nyu,kim2021deformable} is adopted and the network is trained for 20 epochs. 
The NYU v2 dataset consists of 1,449 RGB/D images with feature-aligned color and depth images. 
The low-resolution images are obtained using bicubic downsampling. 
1,000 pairs are used for training and the remaining 449 pairs are used for evaluation. 


Training is conducted using an Adam optimizer with $\beta_1 = 0.9$ and $\beta_2 = 0.999$ and learning rate $0.001$ on a single NVIDIA RTX3090 GPU. 
All networks are efficiently trained end-to-end using \texttt{PyTorch}. 

Network performance will be evaluated quantitatively using the peak signal-to-noise ratio (PSNR) metric and qualitatively using visual comparisons. 
The PSNR can be computed between a generated image $\hat{\mathbf{x}}$ and ground-truth image $\mathbf{x}$ as
\begin{equation}
    \label{eq:psnr}
    \text{PSNR}(\hat{\mathbf{x}},\mathbf{x}) = 20 \log_{10} \left(\frac{\mathbf{x}_\text{max}}{\sqrt{\frac{1}{N} \sum_{i} (\hat{\mathbf{x}}_i - \mathbf{x}_i)^2}} \right),
\end{equation}
where $\mathbf{x}_\text{max}$ is the maximum possible value of the image $\mathbf{x}$ and a higher PSNR (dB) represents a more accurate reconstruction of the ground-truth image. 
Additionally, network inference time, measured as the time required for the network to process a single low-resolution RGB and depth image pair, will be used to compare relative efficiency of each algorithm. 

\subsubsection{Experimental Results}
In this section, we detail the experiments and results using the proposed architectures. 
The networks detailed in Section \ref{sec:methods} are trained using the aforementioned training process for x8 super-resolution yielding the following results. 
Quantitative performance, using the PSNR, is shown in Table \ref{tab:rmse_results} to compare the numerical performance of each algorithm. 

The current state-of-the-art method, DKN, is outperformed by both the \textit{Swin-DMSR} and \textit{NAF-DMSR} proposed architectures for this task. 
Although the computation time is several milliseconds higher than the FDKN \cite{kim2021deformable}, the proposed algorithms achieve superior numerical performance to even the DKN, which requires nearly 5 times the inference time. 
Between the two proposed networks, the \textit{Swin-DMSR} network is able to achieve slightly higher reconstruction performance and \textit{NAF-DMSR} is slightly more computationally efficient. 
Both networks yield state-of-the-art depth map super-resolution performance with competitive efficiency. 
We believe with more fine-tuning, the proposed methods could potentially outperform the FDKN in terms of computation time. 
However, even with the current results, the networks developed in this research achieve state-of-the-art numerical performance. 

A visual comparison of a super-resolution image from the proposed architectures, FDKN, low resolution, and ground-truth images are shown in Fig. \ref{fig:comparison}.
These super-resolution results demonstrate the robustness of our algorithms, which yield superior numerical performance compared to the DKN. 
Furthermore, compared to the network in the project mid-term report, we notice improved high-resolution detail representations in the recovered images by our models while suppressing artifacts present in the FDKN and DKN images, as shown in Fig. \ref{fig:comparison}. 

Through both quantitative and qualitative experimentation, the proposed algorithms outperform the existing state-of-the-art techniques. 

\section{Conclusion}
\label{sec:conclusion}
In this project, we develop novel swin transformer-based \cite{liu2021swin,liang2021swinir} and nonlinear activation free (NAF)-based \cite{chen2022nafnet} algorithms for depth map super-resolution based on joint image filtering achieving state-of-the-art. 
Using these cutting-edge developments in computer vision and image restoration, we develop two novel depth map super-resolution (DMSR) models that leverage the advantages of the joint image filtering technique proposed in \cite{kim2021deformable} in addition to the emerging methods. 
The first network, \textit{Swin-DMSR}, employs a Vision Transformer (ViT) architecture for feature extraction prior to super-resolution reconstruction while the second network, \textit{NAF-DMSR}, uses the NAF approach to reduce computational complexity and improve numerical robustness by replacing nonlinear activation functions with simple multiplication. 
The proposed algorithms are validated in both numerical experimentation and qualitative comparison and demonstrate superior performance to current state-of-the-art DMSR techniques in terms of computational efficiency and reconstruction accuracy. 
Although a trade-off is noted between the \textit{Swin-DMSR} and \textit{NAF-DMSR}, both significantly outperform the DKN and FDKN benchmarks for these two key metrics, enabling DMSR for many crucial use cases that require both low-resolution depth sensors and small platform computationally-constrained hardware. 

{\small
\bibliographystyle{ieee_fullname}
\bibliography{egbib}
}

\end{document}